\documentclass[times,review,10pt]{elsarticle}
\usepackage{longtable}
\usepackage{amsmath}
\usepackage{booktabs}
\usepackage{subcaption}
\usepackage{bbm}
\usepackage{setspace}
\usepackage{xcolor}
\usepackage{environ}
\journal{Pattern Recognition}
\usepackage[breaklinks,hidelinks]{hyperref}

\NewEnviron{commentD}{%
    \color{black}
    \BODY
}

\begin{document}

\begin{highlights}
\item{}%
 Novel training strategy leveraging differentiable clustering for SSL and UDA.
\item{}%
 It improves SSL accuracy by up to 15 percentage points.
\item{}%
 It improves UDA performance of DANN by up to 10 percentage points.
\end{highlights}

\begin{frontmatter}

\title{SuperCM: Improving Semi-Supervised Learning and Domain Adaptation through differentiable clustering\tnoteref{tn1,tn2}}
\tnotetext[tn1]{Visual Intelligence publications are financially supported by the Research Council of Norway, through its Center for Research-based Innovation funding scheme(grant no. 309439), and Consortium Partners}
 
\author[1]{Durgesh Singh\corref{cor1}\fnref{fref1}}
\ead{durgesh.singh@uit.no}
 
\author[2,1]{Ahc\`ene Boubekki\fnref{fref1}}
\ead{ahcene.boubekki@ptb.de}

\author[1,3,4]{Robert Jenssen\fnref{fref1}}
\ead{robert.jenssen@uit.no}

\author[1,3]{Michael Kampffmeyer\fnref{fref1}}
\ead{michael.c.kampffmeyer@uit.no}

\affiliation[1]{organization={Department of Physics and Technology},
 addressline={UiT The Arctic University of Norway}, 
 city={Troms{\o}},
 postcode={9037}, 
 country={Norway}}

\affiliation[2]{organization={Machine Learning and Uncertainty Group},
 addressline={Physikalisch-Technische Bundesanstalt}, 
 city={Berlin},
 postcode={10587}, 
 country={Germany}}
 
\affiliation[3]{organization={Norwegian Computing Center},
 city={Oslo},
 postcode={0373}, 
 country={Norway}}
 
\affiliation[4]{organization={Pioneer Centre for AI},
 city={Copenhagen},
 postcode={1350}, 
 country={Denmark}}

\cortext[cor1]{Corresponding Author}

\tnotetext[tn1]{This work was supported by by the Research Council of Norway~(NFR), through its Center for Research-based Innovation funding scheme~(grant no. 309439), Consortium Partners, and the NFR FRIPRO grant no. 315029.}
\tnotetext[tn2]{Code available at: \url{https://github.com/SFI-Visual-Intelligence/SuperCM-PRJ}}

\fntext[fref1]{UiT Machine Learning Group (machine-learning.uit.no) and with Visual Intelligence, a Norwegian Center for Research-based Innovation (visual-intelligence.no).}

\begin{abstract}
Semi-Supervised Learning (SSL) and Unsupervised Domain Adaptation (UDA) enhance the model performance by exploiting information from labeled and unlabeled data. The clustering assumption has proven advantageous for learning with limited supervision and states that data points belonging to the same cluster in a high-dimensional space should be assigned to the same category. Recent works have utilized different training mechanisms to implicitly enforce this assumption for the SSL and UDA. In this work, we take a different approach by explicitly involving a differentiable clustering module which is extended to leverage the supervised data to compute its centroids. We demonstrate the effectiveness of our straightforward end-to-end training strategy for SSL and UDA over extensive experiments and highlight its benefits, especially in low supervision regimes, both as a standalone model and as a regularizer for existing approaches.
\end{abstract}


\end{frontmatter}

\section{Introduction}\label{sec:introduction}
Deep learning has achieved state-of-the-art performance on various tasks in different domains~\cite{Gu2018}. The success of these models is attributed to supervised learning where large amounts of high-quality labeled data are available. However, obtaining labeled supervision can be difficult in many applications due to expensive and time-consuming annotation processes~\cite{TANG2024110020}. Semi~-Supervised Learning (SSL) and Unsupervised Domain Adaptation (UDA) represent two distinct paradigms in machine learning, which address the challenges associated with learning in scenarios with limited supervision. 

The primary objective of SSL methods is to enhance the performance of the model by utilizing both labeled and unlabeled data. SSL methods depend on the clustering assumption~\cite{Sheikhpour2017141}, which states that data points belonging to the same cluster share common characteristics and leverage unlabeled data to guide the model to recognize similarities over labeled and unlabeled data, resulting in a more robust decision boundary and improved performance. Similarly, the primary objective of UDA~\cite{Liu2022SurveyDA} is to adjust a model trained on the source domain to perform well on a desired target domain where the target domain lacks supervision, i.e. labels are available only for the source domain. Recent UDA approaches have focused on minimizing the covariate shift~\cite{BenDavid2010ATO} between the source and target domain for alignment. However, to perform optimal alignment, the inter-class divergence should be high, and the intra-class divergence between the source and the target domain should be low. Therefore, similar to SSL, UDA can also benefit from the clustering assumption by facilitating the alignment of source and target features that belong to the same cluster. 

Current approaches for SSL and UDA  implicitly attempt to leverage this underlying clustering assumption either through consistency regularization or entropy minimization in the SSL domain and by alternating training schemes in the UDA domain (Section ~\ref{sec:background}). In this study, we instead explore explicit clustering for SSL and UDA in the context of image classification. Recent advances in deep clustering techniques involve simultaneous partitioning of high-dimensional input features through clustering while learning useful representations using deep neural networks~\cite{EZUGWU2022104743}. Capitalizing on the success of these approaches, we introduce a novel training methodology that is inspired by a neural approximation of a Gaussian mixture model~(GMM) and leverages a one-layer auto-encoder named the Clustering Module (CM)~\cite{Boubekki2021JointOO} to enforce the clustering for SSL and UDA. Specifically, we adopted CM for explicit clustering-based regularization which is guided by supervised data points. Our novel training scheme is called Semi-\textbf{Super}vised \textbf{C}lustering \textbf{M}odule (SuperCM) and improves the base SSL and UDA models in an efficient manner with negligible overhead.

The preliminary version of this manuscript previously appeared in~\cite{Singh2023}, concentrating exclusively on SSL and featuring experimental results conducted exclusively on the CIFAR-10 dataset~\cite{cifar10-100}. Here we extend our work by conducting comprehensive experiments encompassing diverse datasets and base models for demonstrating SuperCM's ability in the SSL domain both as a standalone model and as a regularizer for existing SSL approaches. Moreover, we explore the potential benefits of SuperCM regularization for the UDA task and demonstrate that it can easily be integrated into existing UDA approaches.

This paper is structured as follows.
Section~\ref{sec:introduction} describes the objectives of two distinct yet seemingly interconnected tasks, i.e. SSL and UDA, approached from the clustering perspective. 
Section~\ref{sec:background} discusses related work in SSL, UDA, and deep clustering.
Section~\ref{sec:method}, formally provides the definition of SSL \& UDA  and provides a detailed description of our proposed method, i.e. SuperCM, designed to enhance both SSL and UDA performance.
Section~\ref{sec:experiments-ssl}-\ref{sec:analysis-ssl} provides the experimental configurations, results, and analysis of our method regarding the SSL setting.
Similarly, Section~\ref{sec:experiments-da}-\ref{sec:analysis-da} details the experimental setup, results, and further analysis related to the UDA setting. 
Lastly, Section~\ref{sec:conclusion} summarizes our findings, situates our contributions within a broader context, and outlines avenues for future research.

\begin{commentD}
\section{Background}\label{sec:background}
To provide context for SuperCM, prior works in deep semi-supervised learning (SSL)~\cite{Sheikhpour2017141}, unsupervised domain adaptation (UDA)~\cite{Liu2022SurveyDA}, and deep clustering~\cite{EZUGWU2022104743} are reviewed below.

\subsection{Semi-Supervised Learning}\label{subsec:semi-supervised-learning}
Recent SSL methods are typically categorized into consistency-based, entropy-based, and hybrid approaches that incorporate both principles.

\subsubsection{Consistency Regularization}
Consistency regularization is a key idea in semi-supervised learning that assumes a model’s predictions should remain stable under small, label-preserving input perturbations. It encourages the model to produce consistent outputs when exposed to noise or augmentation, helping align decision boundaries with the structure of the input distribution. The $\Pi$-model~\cite{Laine2017TemporalEF} implements this by applying independent augmentations or dropout to the same input and minimizing the difference between the resulting predictions. Mean Teacher~\cite{Tarvainen2017} extends this using a teacher-student framework, where the teacher’s weights are an exponential moving average of the student’s, the student matches the teacher’s output under different augmentations. Virtual Adversarial Training (VAT)~\cite{Miyato2019VirtualAT} takes a more targeted approach by computing the perturbation that most changes the model’s prediction and enforcing consistency against it. This promotes smoothness in high-sensitivity regions and leads to stronger generalization near decision boundaries.

\subsubsection{Entropy Minimization}
Entropy minimization encourages confident predictions on unlabeled data~\cite{Grandvalet2005}, often implemented through pseudo-labeling~\cite{Dong-HyunLee2013}, though its effectiveness can suffer from noisy or miscalibrated predictions. To mitigate this, DMT~\cite{FENG2022108777Pat} employs dual models with disagreement-based filtering to suppress unreliable pseudo labels, while Semi-PCT~\cite{Wang2022} uses progressive cross-training to gradually refine predictions across iterations. PL-Mix~\cite{wang2025plmix} refines pseudo labels through iterative mixup, guided by consistency between weak and strong augmentations to improve label reliability. Furthermore, S2Match~\cite{guan2025s2match} incorporates self-paced learning and confidence-aware sampling to filter out uncertain pseudo labels during training. Finally, BFA~\cite{huang2025boundary} enhances semantic segmentation by aligning structural and boundary-level features between labeled and unlabeled data, leading to better spatial consistency.

\subsubsection{Hybrid Approaches}
Hybrid methods in semi-supervised learning integrate consistency regularization with entropy-based objectives to leverage the strengths of both paradigms. A prominent example is ICT (Interpolation Consistency Training)~\cite{Verma2019ICT}, which introduced the idea of enforcing consistency not just on individual samples, but also on interpolated inputs and their corresponding pseudo-labels, encouraging the model to behave linearly between data points. Building on this, MixMatch~\cite{Berthelot2019MixMatch} proposed a unified framework that combines multiple components: label guessing using model predictions, MixUp~\cite{Zhang2018mixup} for interpolating both inputs and labels, and an entropy minimization loss to encourage confident predictions. FixMatch~\cite{Sohn2020FixMatch} further streamlined this approach by applying a simple yet effective strategy: generate pseudo-labels from weakly augmented inputs, and enforce consistency on corresponding strongly augmented versions, but only when the model is confident. SequenceMatch~\cite{Nguyen2023SequenceMatch} refined this pipeline by introducing a sequence of intermediate-strength augmentations and applying filtering mechanisms to reduce noise in pseudo-labels, resulting in smoother learning dynamics. More recently, GraphixMatch~\cite{Xu_2024_GraphixMatch} extended the hybrid consistency-entropy framework to graph-structured data, incorporating domain-specific graph adaptation to better capture relational dependencies and structure-aware consistency.

\subsection{Unsupervised Domain Adaptation}
UDA methods aim to mitigate distribution shifts between source and target domains~\cite{Liu2022SurveyDA}. They are generally categorized into discrepancy-based and adversarial approaches, both of which provide complementary mechanisms for domain alignment.

\subsubsection{Discrepancy-Based Domain Adaptation}
In discrepancy-based UDA, domain alignment is achieved by minimizing distributional differences. Maximum Mean Discrepancy (MMD)~\cite{Gretton2012MMD} has been widely adopted to quantify divergence between source and target means in the feature space. Structured extensions such as DJSA~\cite{Cao2024DJSA} have been proposed to align intra- and inter-class subdomains for improved semantic transfer. Meanwhile, regularization-based strategies such as MCC~\cite{MCCLoss2020} and BNM~\cite{BNM2020} have been applied to encourage low-confidence suppression and discriminative feature learning respectively.

\subsubsection{Adversarial Domain Adaptation}
Adversarial domain adaptation seeks to learn domain-invariant features by jointly training a feature extractor and a domain discriminator in a minimax setup. DANN~\cite{Ganin2015DANN} formalized this framework using a gradient reversal layer to align source and target distributions while preserving task-relevant information. A black-box, source-free approach BPDA~\cite{shi2023bpda} later extended this idea via distributionally adversarial training with third-party supervision, enhancing privacy and generalization. Building on adversarial alignment, PLADA~\cite{fang2024plada} introduces prototype learning and a weighted prototype loss to align both global and category-level distributions for more discriminative representations.

\subsection{Deep Clustering}
Deep clustering techniques~\cite{EZUGWU2022104743} aim to simultaneously learn feature representations and cluster assignments. Earlier works~\cite{Caron2018DeepCF, Li2021PrototypicalCL} followed iterative schemes involving alternate optimization of cluster labels and representations. More recent methods employ end-to-end training for cluster-friendly embeddings. A notable example is the Clustering Module (CM)~\cite{Boubekki2021JointOO}, which integrates a shallow autoencoder and a deep feature extractor, using GMM-based losses to drive clustering. The CM will be further summarized in Section~\ref{subsubsec:cm} and extended to SSL and UDA scenarios in Section~\ref{subsec:supercm_ssl}.
\end{commentD}
 
\subsubsection{Clustering Module~(CM)}\label{subsubsec:cm}

At the core of our approach lies the CM, originally proposed in~\cite{Boubekki2021JointOO}, which integrates the clustering objective directly into the training of an autoencoder~(AE) backbone. Unlike traditional methods that treat representation learning and clustering as separate stages, the CM jointly optimizes a reparameterization of the $\mathcal{Q}$-function from the Expectation-Maximization (EM) algorithm for GMMs.

Given a dataset of $n$ input samples, each passed through the encoder part of the AE to obtain latent representations ${\pmb x}_i \in \mathbb{R}^d$, assumed to be drawn from one of $K$ latent isotropic clusters, the CM introduces a differentiable objective that encourages the learning of meaningful cluster assignments. The loss function of the CM is defined as:

\begin{multline}
\label{eq:qfunc}
\mathcal{L}_{\text{CM}} = \frac{1}{n} \Bigg( 
\underbrace{-\sum_{i=1}^n \| {\pmb x}_i - \bar{{\pmb x}}_i \|^2}_{E_1} 
- \underbrace{\sum_{i=1}^n \sum_{k=1}^K \gamma_{ik}(1 - \gamma_{ik}) \| {\pmb \mu}_k \|^2}_{E_2} \\
+ \underbrace{\sum_{i=1}^n \sum_{\substack{j,l=1 \\ j \ne l}}^K \gamma_{ij} \gamma_{il} {\pmb \mu}_j^\top {\pmb \mu}_l}_{E_3} 
- \underbrace{\sum_{k=1}^K (1 - \alpha) \log \tilde{\gamma}_k}_{E_4} 
\Bigg)
\end{multline}

\textbf{Reconstruction Loss~(E1):}  
This term ensures that the latent representation preserves the essential information needed to accurately reconstruct the input. Here, ${\pmb x}_i$ denotes the output of the encoder (i.e., the latent embedding), and $\bar{{\pmb x}}_i$ is the reconstruction produced by the CM.

\textbf{Sparsity and cluster merging~(E2 \& E3):}  
In the two-cluster case, where the soft assignment for sample \(i\) to cluster 1 is denoted by \(\gamma_i \in [0, 1]\), and the centroids are \(\mu_1\) and \(\mu_2\), the combined effect of the \(E_2\) and the \(E_3\) simplifies to:

\begin{equation}
\label{eq:e2e3_two_cluster}
E_2 + E_3 = \gamma_i(1 - \gamma_i) \left( \|\mu_1\|^2 + \|\mu_2\|^2 - 2 \mu_1^\top \mu_2 \right)
\end{equation}

This expression is weighted by the term \(\gamma_i(1 - \gamma_i)\), which measures the uncertainty of the sample’s cluster assignment. This value is close to zero when the assignment is confident (i.e., \(\gamma_i \approx 0\) or \(1\)), and peaks at \(\gamma_i = 0.5\), where the model is maximally uncertain.
Minimizing \(E_2 + E_3\) thus has two simultaneous effects: (i) it encourages \emph{sparse and confident cluster assignments}, because high uncertainty increases the penalty; and (ii) it promotes \emph{cluster merging} by pulling the centroids \(\mu_1\) and \(\mu_2\) closer together when assignments are ambiguous, leading to a more compact clustering representation.

\textbf{Cluster Prior Regularization (E4):}  
This term imposes a Dirichlet prior on cluster usage, where \(\tilde{\gamma}_k = \frac{1}{n} \sum_{i=1}^n \gamma_{ik}\) denotes the average responsibility for cluster \(k\). When \(\alpha > 1\), the regularization encourages balanced cluster assignments and prevents dominance by a few clusters.

The CM consists of an encoder \& decoder, where the encoder produces soft assignments $\Gamma = \{ \gamma_{ik} = P(y_i = k \mid {\pmb x}_i) \}$ using a softmax activation, with each $\gamma_{ik}$ representing the posterior probability of assigning sample $i$ to cluster $k$. The centroids ${\pmb \mu}_k \in \mathbb{R}^d$ are learnable parameters of the decoder, serving as representatives of the cluster structure in the latent space. This formulation allows the CM to jointly learn compact and informative embeddings alongside discriminative clusters in an end-to-end differentiable way, explicitly integrating clustering-based regularization into the training of the parent AE backbone.

\section{Method}\label{sec3}
\label{sec:method}

\subsection{Problem Definition}
We now provide a formal description of the SSL and UDA problems for the classification of images in the $K$ classes.

\subsubsection{SSL}
Given a training dataset $\mathcal{D}=\mathcal{D}_{l}\cup\mathcal{D}_{u}$ made up of a set of labeled images $\mathcal{D}_{l}=\{\mathcal{I}^{(l)},\mathcal{Y}^{(l)}\}$ and of a set of unlabeled images $\mathcal{D}_{u}=\{\mathcal{I}^{(u)}\}$ belonging to one of the $K$ classes. The two subsets are of cardinality $N_l$ and $N_u$, respectively. The predictive function of a supervised-only model trained on labeled data is defined as $\emph{p}(\mathcal{Y}^{(l)}|\mathcal{I}^{(l)})$. The objective of an SSL algorithm is to improve $\emph{p}(\mathcal{Y}^{(l)}|\mathcal{I}^{(l)} \cup \mathcal{I}^{(u)})$ by making optimal use of the labeled and unlabeled training sets. By leveraging information from labeled and unlabeled data SSL algorithms strive to improve a model's generalization capability on the previously unseen data, when compared to its supervised-only baseline. 

\subsubsection{UDA}
Given a training dataset $\mathcal{D}=\mathcal{D}_{s}\cup\mathcal{D}_{t}$, made of a source data set $\mathcal{D}_{s}=\{\mathcal{I}^{(s)},\mathcal{Y}^{(s)}\}$ and an unlabeled target data set $\mathcal{D}_{t}=\{\mathcal{I}^{(t)}\}$, which contain each $N_s$ and $N_t$ elements, respectively. 
The goal of UDA algorithms is to fine-tune a model trained on the source data to minimize the domain shift with the target data, that is, to improve its predictions on the target domain, $\emph{p}_t(\mathcal{Y}^{(t)}|\mathcal{I}^{(t)})$.

\subsection{SuperCM for improving SSL and UDA} 
\label{subsec:supercm_ssl}
Given an SSL/UDA algorithm with an SSL base model for image classification into $K$ classes. Our approach appends the CM module to the backbone feature extractor and leverages the supervised data through the CE loss over the posterior probabilities along with the $\mathcal{L}_{\rm CM}$ loss. However, we observe that this could lead to trivial solutions when the training of the backbone is too fast.
Since the model has access to some labels, we learn the centroids as a class-wise average of the labeled source domain data instead of learning them with gradient descent. Specifically, we rely on a moving average to prevent frequent noisy updates. 
Computing the centroids this way also prevents them from collapsing and constrains the cluster probabilities, making the Dirichlet prior redundant. We therefore discard it by setting the value of the Dirichlet prior coefficient $\alpha=1$ in Equation 1. The performance of the base SSL and UDA models is improved by employing this straightforward strategy, all without relying on any further complex training mechanism.

At each iteration $\tau$, the input batch $B$ consists of $n_B^{(l)}$ labeled data pairs $\{\mathcal{I}_B^{(l)},\mathcal{Y}_B^{(l)}\}$ and $n_B^{(u)}$ unlabeled data $\mathcal{I}_B^{(u)}$.
Both inputs are first transformed by a feature extractor $F$ into $\mathcal{X}_B^{(l)}=F\big(\mathcal{I}_B^{(l)}\big)$ and $\mathcal{X}_B^{(u)}=F\big(\mathcal{I}_B^{(u)}\big)$.
The centroids are then updated using the labeled data as follows:

\begin{equation}
    {\pmb \mu}_{k} = \frac{\tau-1}{\tau} {\pmb \mu}_{k} + \frac{1}{\tau} \frac{1}{n_B^{(l)}} \sum_{i=1}^{n_B^{(l)}} \mathbbm{1}_{(y^{(l)}_{i}=k)} {\pmb x}^{(l)}_i
\label{eq:setcen}
\end{equation}

Finally, the labeled and unlabeled data are concatenated and passed through the CM to obtain the classification probabilities $\Gamma_B^{(l+u)}$ as well as the reconstructions $\bar{\mathcal{X}}_B^{l+u}$.

The loss function of the SuperCM combines a standard CE loss applied to the classification responsibilities of the labeled data and the CM loss applied to both types of data. SuperCM may also be used as a regularizer for existing SSL and UDA models in which case the combined loss can be stated as follows:

\begin{multline}
    \mathcal{L}^{\rm{SSL/UDA}}_{\rm SuperCM} = \mathbf{CE}( \Gamma_B^{(l)}, \mathcal{Y}_B^{(l)} ) +  
              \beta \cdot \mathcal{L}_{\rm CM}( \mathcal{X}_B^{(l+u)}, \Gamma_B^{(l+u)}, \bar{\mathcal{X}}_B^{(l+u)}) 
            + \delta \cdot \mathcal{L}_{\rm (SSL/UDA)}
\label{eq:supcm}
\end{multline}

\begin{figure}[!t]
\centering
\includegraphics[width=\textwidth]{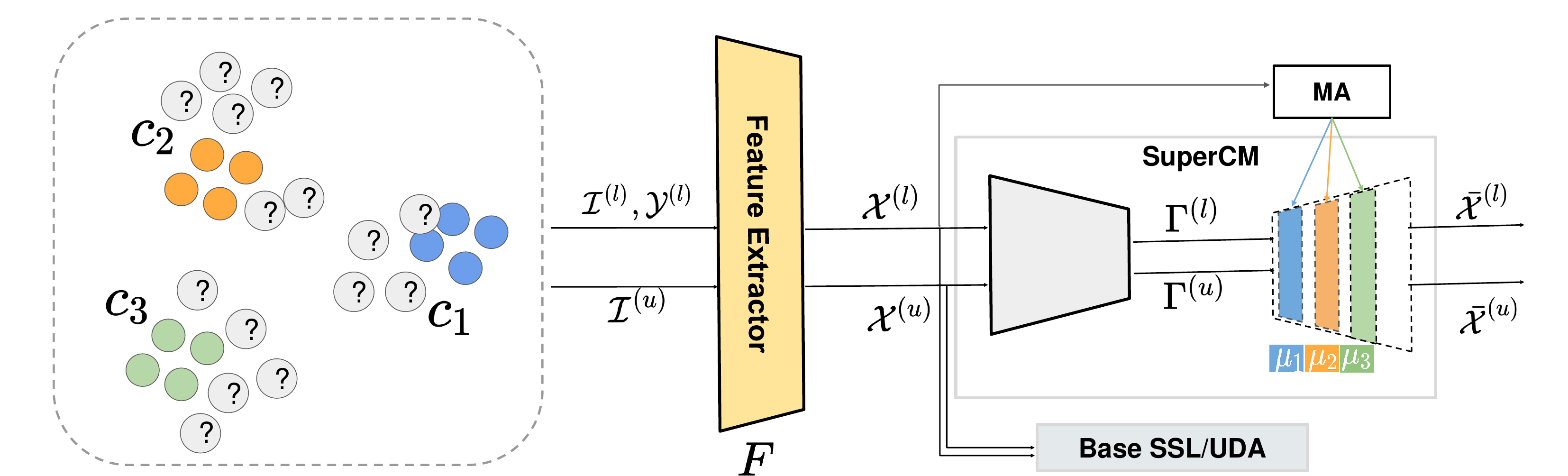}
\caption{Architecture of our proposed method}
\label{fig:supercm_arch}
\end{figure}

Where $\beta \geq 0$ and $\delta \geq 0$ weights the CM loss and the loss of the base SSL/UDA  model, respectively. The architecture of the SuperCM is provided in Figure \ref{fig:supercm_arch}.

\section{Semi-Supervised Learning Experiments}\label{sec:experiments-ssl}
In this section, we provide details for the SSL experiments. Specifically, we performed two types of experiments: (i)~evaluating SuperCM as an SSL approach, and (ii)~using it as a regularizer for five existing approaches: $\Pi$-Model~\cite{Laine2017TemporalEF}, Mean-Teacher~\cite{Tarvainen2017}, VAT~\cite{Miyato2019VirtualAT}, Pseudo-labels~\cite{Dong-HyunLee2013} and ICT~\cite{Verma2019ICT}.

\subsection{Dataset}
 
We conducted experiments on a diverse range of datasets to thoroughly analyze SuperCM's performance. We now provide a description for each of the public datasets.

\paragraph{\textbf{MNIST}}
The MNIST~\cite{mnist} dataset comprises grayscale images of handwritten digits (0 to 9), organized into 10 classes. The images are pre-processed to 28x28 pixels. With 60000 training images and 10000 testing images, MNIST is a low-complexity dataset. In the SSL scenario, the training set is split into a labeled set and an unlabeled set, allowing SSL algorithms to benchmark their performance using both labeled and unlabeled data.

\paragraph{\textbf{SVHN}}
The SVHN (Street View House Numbers) dataset features digits (0 to 9) in color images sourced from Google Street View. Cropped to 32x32 pixels, the dataset is categorized into 10 classes. It includes a training set with 73257 images, a test set with 26032 images, and an extra dataset with 130068 images. SVHN is a crucial benchmark for SSL algorithms due to its substantial number of unlabeled images. 

\paragraph{\textbf{CIFAR-10 and CIFAR-100}}
CIFAR-10~\cite{cifar10-100} and CIFAR-100~\cite{cifar10-100} datasets contain 60000 32x32 pixel color images. Both datasets represent real-world objects, with a balanced split of 50000 training and 10000 testing images. The CIFAR-10 dataset consists of 10 classes, while CIFAR-100 consists of 100 classes. These datasets serve as challenging benchmarks for image classification tasks, evaluating SSL algorithms under varying levels of supervision. CIFAR-100, with its finer-grained classes, presents a more intricate challenge for algorithms to discern subtle differences between categories.

\paragraph{\textbf{STL-10}}
STL-10~\cite{stl10} is inspired by the CIFAR-10 dataset and specifically designed for SSL and Transfer Learning tasks. It consists of 96x96-pixel color images, making it more complex and detailed than CIFAR-10. STL-10 consists of a total of 5000 labeled training images, 8000 labeled test images, and  100000 unlabeled images.

\subsection{Architecture}

We now present the model architecture utilized for the  SSL task. In the SSL scenario, we follow~\cite{Oliver2018}, for conducting experiments with SuperCM. For CIFAR-10, SVHN, and MNIST datasets, we employ a Wide-Resnet-28-2 backbone, while for CIFAR-100 and STL-10 datasets, we use Wide-Resnet-28-8. The Wide-ResNet-28-2 comprises 28 layers with a widening factor of 2, resulting in approximately 1.5 million parameters and generating a pre-classifier feature vector of dimension 128. In contrast, Wide-ResNet-28-8 consists of 28 layers with a widening factor of 8, yielding around 36.5 million parameters and feature vectors of dimension 512.

\subsection{Experimental Settings}
This section presents implementation details of SSL experiments, covering aspects such as data augmentation, model training, baseline methods, and evaluation procedures. We provide a detailed description of each aspect in the subsequent subsections.

\subsubsection{Data Preprocessing}
We split all the training datasets into a 90-10 split for model training and validation, with the validation split also being used for hyperparameter tuning. A small proportion of the training split is selected as labeled data, and the rest is considered unlabeled for the SSL training. Varying supervision levels are detailed in Section~\ref{sec:results-ssl}. The test split is then used for reporting model performance. Moreover, we use augmentations flip, random crop, and Gaussian noise to prevent overfitting for CIFAR-10, CIFAR-100, and STL-10. We use random crop augmentation for MNIST and SVHN datasets.

\subsubsection{Baselines}
\label{subsec:baselines}
We consider the performance when only optimizing the CE loss as our supervised-only baseline, that is no base SSL model is used i.e. $\delta=0$ and $\beta=0$ in Eq.~\eqref{eq:supcm}. This baseline helps us to evaluate the standalone performance of SuperCM for the SSL task.
Furthermore, we utilize SuperCM training as an unsupervised regularizer for different base SSL models namely  $\Pi$-Model, Mean-Teacher, VAT, Pseudo-labels, and ICT. We consider the base model performance i.e. $\delta\neq0$ and $\beta=0$ in Eq.~\eqref{eq:supcm} as the baseline for these experiments.

\subsubsection{Model Training}
We follow the recommendations of~\cite{Oliver2018} regarding data preprocessing and training protocols for all SSL methods. All models were trained for 500000 iterations with a batch size of 100. For experiments involving the $\Pi$-Model, Mean Teacher, VAT, and Pseudo-Label methods, the Adam optimizer~\cite{Kingma2015AdamAM} was used with a step-based learning rate scheduler, decaying the learning rate by a factor of 0.1 after 400000 iterations. For the ICT model, we used stochastic gradient descent with Nesterov momentum~\cite{Nesterov1983AMF} (momentum = 0.9), combined with cosine annealing~\cite{Loshchilov2016SGDRSG} starting from an initial learning rate of 0.1. We also applied L2 regularization with a weight decay coefficient of 0.0001.

Additionally, the hyperparameters $\beta$ and $\delta$ were tuned using Bayesian optimization via the Weights \& Biases platform~\cite{wandb}. The search space for both parameters was defined over a logarithmic scale ranging from $0$ to $10$. Tuning was guided by validation performance, and the optimizer adaptively explored the space to capture potential nonlinear effects. Table~\ref{tab:gamma_beta_values} summarizes the final values of $10^{\delta}$ and $10^{\beta}$ used for each method.

\begin{table}[h]
\caption{Hyperparameter values of $10^{\delta}$ and $10^{\beta}$ for training with different base SSL models and the SuperCM module.}
\centering
\begin{tabular}{lcc}
\toprule
\textbf{base SSL model} & $10^{\delta}$ & $10^{\beta}$ \\
\midrule
None & 0.00 & 0.18 \\
MT~\cite{Tarvainen2017}   & 0.15 & 0.12 \\
VAT~\cite{Miyato2019VirtualAT}  & 5.62 & 0.63 \\
PI~\cite{Laine2017TemporalEF}   & 0.19 & 0.17 \\
PL~\cite{Grandvalet2005}   & 0.18 & 3.01 \\
ICT~\cite{Verma2019ICT}  & 0.16 & 2.23 \\
\bottomrule
\end{tabular}
\label{tab:gamma_beta_values}
\end{table}

\subsubsection{Evaluation Metrics}
We evaluate the best Top-1 accuracy on the validation set for model selection. Overall model performance is measured using Top-1 accuracy on the test set.
We report mean and standard deviation over three runs, trained with different random seeds. Additionally, we follow~\cite{Oliver2018} and compute the final model based on stochastic weight averaging and report the score of the final model to compare the performance of CE and SuperCM.

\setlength\LTleft{0cm}
\setlength\LTright{0cm}
\setlength\LTpre{1cm}
\setlength\LTcapwidth{\linewidth}

\begin{longtable}{@{\extracolsep{\fill}}lcccc}
\multicolumn{5}{r}{  Continued ...} \endfoot \\
\multicolumn{5}{r}{} \endlastfoot
    \caption{Results for Top-1 accuracy with CE and  SuperCM for different base models on various SSL datasets and levels of supervision. 
    None denotes the SSL setting without a base model where performance of CE and SuperCM is compared. Bold numbers indicate statistically significant improvements (t-test, $p \leq 0.05$).} \\ \toprule 
                                             &   CE            &   SuperCM                &       CE        &      SuperCM            \\  \midrule \midrule
    
        & \multicolumn{4}{c}{MNIST~\cite{mnist}} \\
    
    & \multicolumn{2}{c}{ 100 labels} 		        & \multicolumn{2}{c}{ 1000 labels}                                       \\
    None                                     & 82.28$\pm$0.95  & \textbf{97.45$\pm$0.30}  & 98.30$\pm$0.15  & \textbf{99.24$\pm$0.11} \\  
    Pseudo-Label~\cite{Dong-HyunLee2013}     & 95.55$\pm$0.43  & \textbf{96.86$\pm$0.19}  & 99.07$\pm$0.14  & \textbf{99.16$\pm$0.20} \\
    VAT~\cite{Miyato2019VirtualAT}           & 97.14$\pm$0.38  & \textbf{98.84$\pm$0.06}  & 99.18$\pm$0.14  & \textbf{99.29$\pm$0.08} \\
    $\Pi$-Model~\cite{Laine2017TemporalEF}   & 93.68$\pm$0.54  & \textbf{97.46$\pm$0.94}  & 98.88$\pm$0.04  & \textbf{99.26$\pm$0.10} \\
    Mean Teacher~\cite{Tarvainen2017}        & 96.32$\pm$0.19  &          96.82$\pm$0.55  & 98.91$\pm$0.14  &          99.36$\pm$0.05 \\ \midrule
    
    & \multicolumn{4}{c}{SVHN~\cite{svhn}}		              \\
    
    & \multicolumn{2}{c}{250 labels}		        & \multicolumn{2}{c}{1000 labels}                                        \\
    None                                     & 70.56$\pm$1.06  & \textbf{85.69$\pm$2.25}  & 87.43$\pm$0.29  & \textbf{92.59$\pm$0.15} \\  
    Pseudo-Label~\cite{Dong-HyunLee2013}     & 81.15$\pm$5.36  &          79.87$\pm$1.55  & 92.77$\pm$0.65  &          93.39$\pm$0.47 \\
    VAT~\cite{Miyato2019VirtualAT}           & 76.70$\pm$1.30  &          78.19$\pm$3.10  & 83.23$\pm$1.75  &          86.60$\pm$1.00 \\
    $\Pi$-Model~\cite{Laine2017TemporalEF}   & 82.37$\pm$0.56  & \textbf{87.41$\pm$0.84}  & 91.88$\pm$0.43  & \textbf{93.83$\pm$0.31} \\
    Mean Teacher~\cite{Tarvainen2017}        & 71.07$\pm$2.32  &          75.52$\pm$3.48  & 88.02$\pm$0.40  &           \textbf{90.72$\pm$0.23} \\ \midrule
    
    & \multicolumn{4}{c}{CIFAR-10~\cite{cifar10-100}}  \\
    
    & \multicolumn{2}{c}{600 labels}                    & \multicolumn{2}{c}{4000 labels}                           \\
  
    None                                     & 56.94$\pm$0.46  & \textbf{62.14$\pm$1.40}  & 78.65$\pm$0.45  & \textbf{82.26$\pm$0.26} \\  
    Pseudo-Label~\cite{Dong-HyunLee2013}     & 61.05$\pm$1.25  &          65.19$\pm$2.52  & 84.97$\pm$0.22  &          85.19$\pm$0.47 \\
    VAT~\cite{Miyato2019VirtualAT}           & 68.43$\pm$0.89  & \textbf{75.23$\pm$3.92}  & 86.82$\pm$0.19  &          86.69$\pm$0.11 \\
    $\Pi$-Model~\cite{Laine2017TemporalEF}   & 58.67$\pm$1.35  & \textbf{64.22$\pm$1.24}  & 84.08$\pm$0.45  & \textbf{84.94$\pm$0.13} \\
    Mean Teacher~\cite{Tarvainen2017}        & 59.82$\pm$0.76  &          64.11$\pm$2.65  & 84.53$\pm$0.35  &          84.64$\pm$0.37 \\ \midrule
     & \multicolumn{4}{c}{CIFAR-100~\cite{cifar10-100}} \\
     
    & \multicolumn{2}{c}{ 2500 labels}         & \multicolumn{2}{c}{10000 labels}                                  \\ 
    None                                     & 33.57$\pm$0.20  &          34.05$\pm$2.25  & 55.47$\pm$0.18  &          56.24$\pm$1.09 \\  
    VAT~\cite{Miyato2019VirtualAT}           & 37.57$\pm$0.64  & \textbf{43.55$\pm$0.55}  & 60.12$\pm$0.26  & \textbf{60.47$\pm$0.09} \\
    ICT~\cite{Verma2019ICT}                  & 50.26$\pm$0.32  & \textbf{51.15$\pm$0.36}  & 66.06$\pm$0.44  &          66.01$\pm$0.23 \\  \midrule

    & \multicolumn{4}{c}{STL-10~\cite{stl10}} \\
    
    & \multicolumn{2}{c}{600 labels} 		    & \multicolumn{2}{c}{4000 labels}                                      \\
    None                                     & 59.94$\pm$1.15  &          61.16$\pm$2.48  & 83.09$\pm$0.42  &          82.50$\pm$0.67 \\  
    Pseudo-Label~\cite{Dong-HyunLee2013}     & 61.65$\pm$0.63  &          55.21$\pm$4.29  & 81.77$\pm$0.54  &          81.18$\pm$0.97 \\
    VAT~\cite{Miyato2019VirtualAT}           & 60.11$\pm$1.09  &          60.10$\pm$1.37  & 80.91$\pm$0.63  &          82.58$\pm$0.67 \\
    $\Pi$-Model~\cite{Laine2017TemporalEF}   & 60.59$\pm$0.72  & \textbf{62.94$\pm$0.21}  & 82.12$\pm$0.33  & \textbf{82.74$\pm$0.53} \\
    Mean Teacher~\cite{Tarvainen2017}        & 61.27$\pm$0.39  &          62.14$\pm$0.47  & 82.16$\pm$0.76  &          83.24$\pm$0.92 \\  \bottomrule

     \label{tab:results-ssl}
  
\end{longtable}

\section{ SSL Results}
\label{sec:results-ssl}
Table~\ref{tab:results-ssl}  summarizes the results for the SSL experiments. We briefly discuss the experiments on each of the previously mentioned datasets for different levels of supervision.

\paragraph{\textbf{MNIST}}
In the MNIST experiments with 100 and 1000 labels, SuperCM without a base model improves accuracy by $15.17\%$ and $0.94\%$, respectively. As a regularizer, SuperCM enhances performance for Pseudo-label ($1.31\%$), VAT ($1.70\%$), $\Pi$-model ($3.78\%$), and Mean Teacher model ($0.50\%$) in the 100 label setting. In the 1000 label setting, improvements are observed for Pseudo-label ($0.09\%$), VAT ($0.11\%$), $\Pi$-model ($0.38\%$), and the Mean Teacher model ($0.45\%$).  Overall we observe that SuperCM training for SSL and regularization consistently outperforms CE training for the MNIST dataset. 

\paragraph{\textbf{SVHN}}
For the SVHN dataset, we observe that SuperCM alone achieves performance improvement of $15.13\%$ and $5.16\%$ for 250 and 1000 label settings for the SSL task. When SuperCM is used as a regularizer for VAT, $\Pi$-Model, and Mean teacher models the improvements reach $1.49\%$, $5.04\%$, and $4.45\%$ for the 250 label setting and $3.37\%$, $1.95\%$, and $2.70\%$ for the 1000 label setting. These experiments conclude that SuperCM training demonstrates effectiveness in improving the performance of most SSL scenarios on the SVHN dataset.

\paragraph{\textbf{CIFAR-10}}
Without SSL base model, i.e., $\delta=0$, SuperCM significantly improves the performance over the CE baseline by $5.2\%$ and $3.6\%$ on the 600 and 4000 labels setting, respectively.
When SuperCM is used as a regularizer of a base model, the accuracy of the base model significantly increases for the 600-label setting.
Specifically, SuperCM improves the accuracy of the Pseudo-Label baseline by $4.14\%$ and that of VAT by $6.8\%$.
However, we do not observe significant improvement for the 4000 labels setting. 

\paragraph{\textbf{CIFAR-100}}
On the more complex CIFAR-100 dataset with 100 classes, our results showcase noteworthy improvements. When SuperCM operates independently (SSL base model is None), there is a marginal enhancement of $0.48\%$ and $0.77\%$ for settings with 2500 and 10000 labels, respectively. Introducing SuperCM to the ICT base model leads to a slight decrease of $0.05\%$ with 10000 labels but brings about an increase of $0.89\%$ with 2500 labels. In the case of using VAT as the base SSL model, the SuperCM yields a $0.35\%$ improvement for the 10000 label setting, while a substantial improvement of $5.98\%$ is observed for the 2500 label setting.

\paragraph{\textbf{STL-10}}
For the STL-10 dataset, we conduct experiments using 400 and 4000 labeled data from the training set. We observe in Table~\ref{tab:results-ssl} that for the 600 label setting, SuperCM achieves performance improvement of $1.22\%$ as compared to the CE training. Similarly, performance improvement of $2.35\%$ and  $0.87\%$ was observed with $\Pi$-Model and Mean Teacher base models. With 4000 labels, we observe marginal improvements of SuperCM over CE training where, SuperCM achieved performance improvement of $0.62\%$, $1.08\%$ and $1.67\%$ with respect to $\Pi$-Model, Mean Teacher and VAT and performance of  Pseudo-Label decreased by $0.59\%$ over CE.

\paragraph{\textbf{Summary}}
We hypothesize that in the low supervision setting the SSL base models benefit from the well-separated clustering obtained by SuperCM when it is difficult to obtain a reliable supervisory signal from the CE loss during the training. A visualization of the well-separated features learned by SuperCM is provided in Sec.~\ref{sec:featureVis} for different datasets.

\section{SSL Analysis}

\subsection{Amount of Supervision}
\begin{figure}[!t]
\centering
\includegraphics[width=0.8\textwidth]{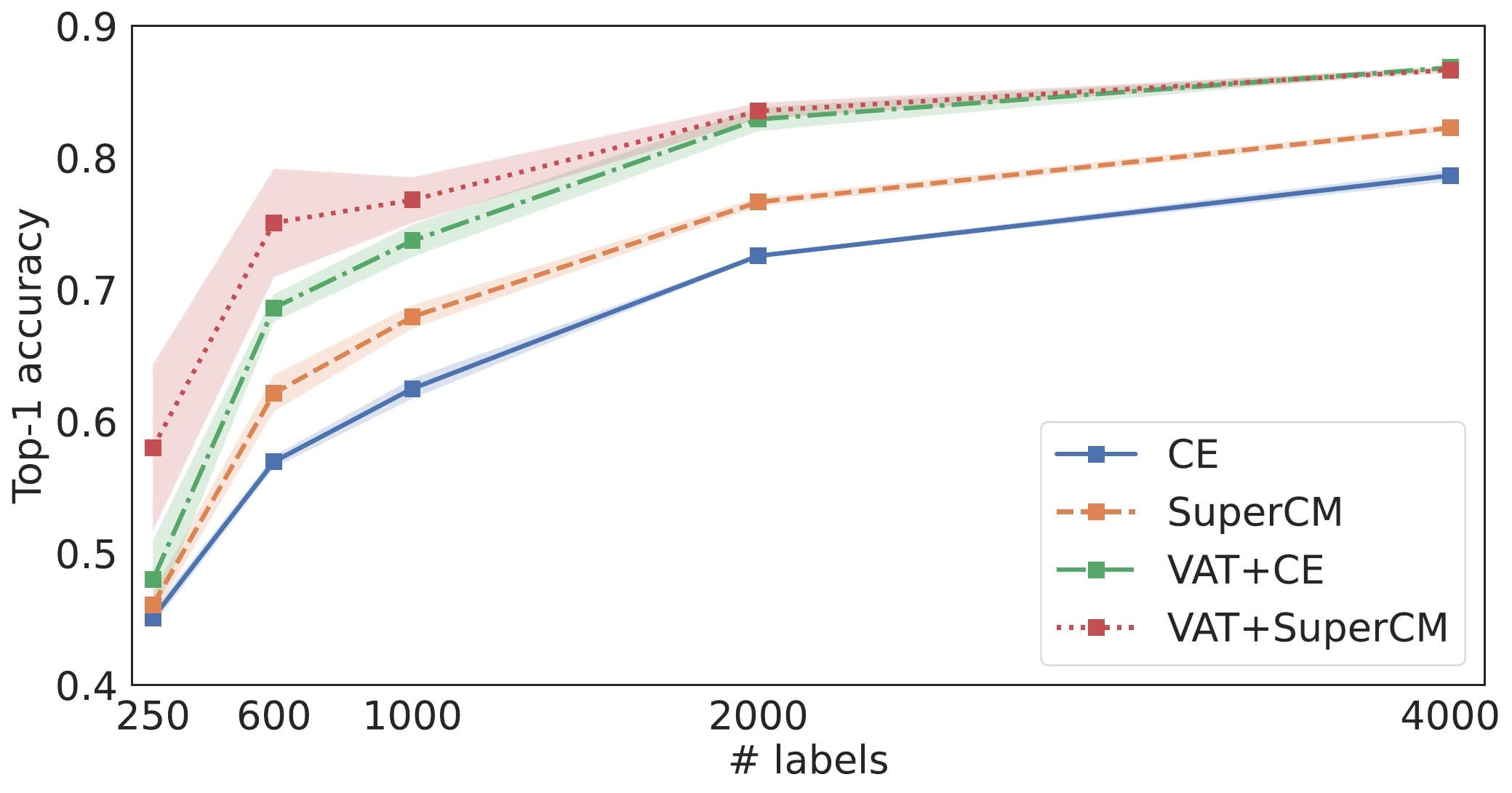}
\caption{Training with CE and SuperCM for different supervision on CIFAR-10~\cite{cifar10-100} dataset.}
\label{fig:vary}
\end{figure}

We evaluate the performance of our method on CIFAR-10 for different training amounts of label data ranging from 250 to 4000 instances, with and without  VAT as the SSL base model. The results are summarized in Figure~\ref{fig:vary}.
Without the SSL base model, SuperCM improves the supervised-only baseline (CE) for all levels of supervision. However, we observe that the difference is smaller in the smallest setting with 250 labels. In this case, we do not expect a large performance improvement from SuperCM alone, as the supervision is extremely scarce and not sufficient to learn a cluster-friendly embedding for the overarching SSL task. However, as the number of labels increases, we see significant improvements compared to the supervised-only baseline ranging from $3\%$ to $6\%$.
Similarly, using SuperCM as a regularizer improves the performance of VAT significantly until the 2000 setting. The SuperCM regularization improves the VAT baseline by around $10\%$ in the 250-label setting. We also see performance improvements of around $6\%$  and $3\%$ in the case of the 600 and 1000 label settings.

\subsection{Hyper-parameter Sensitivity}
\begin{figure}[!t]
\centering
\includegraphics[width=0.8\textwidth]{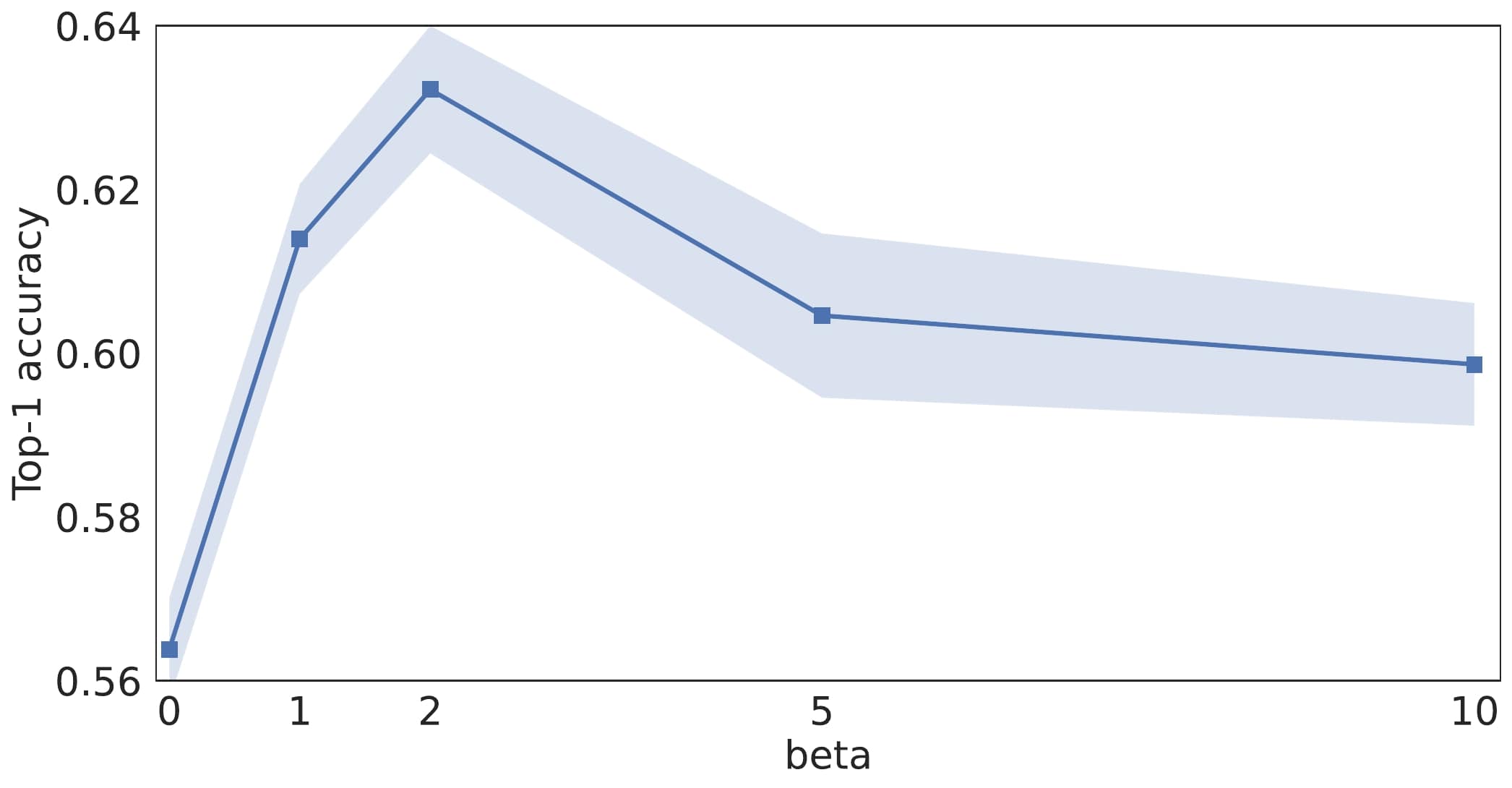}
\caption{Top-1 accuracy of SuperCM trained with 600 labels and different values of the hyper-parameter $\beta$.}
\label{fig:paramsen}
\end{figure}

Without a base model, the SuperCM uses a single hyper-parameter $\beta$ controlling the influence of the clustering loss. 
To study the sensitivity of the model toward $\beta$, we trained SuperCM with 600 labels and varying values of $\beta$ on the CIFAR-10 dataset. Figure~\ref{fig:paramsen} shows the Top-1 accuracy of the model trained with different values of $\beta$, where $\beta=0$ represents the training with only supervised loss (CE). As the influence of the CM grows ($\beta >0$), we observe considerable performance improvement. However, we further observe that as the weight for the CM increases and less weight is given to the CE, performance slowly decreases.

\subsection{Feature Visualization}\label{sec:featureVis}

\begin{figure*}[t]
   \centering
   \includegraphics[width=\textwidth]{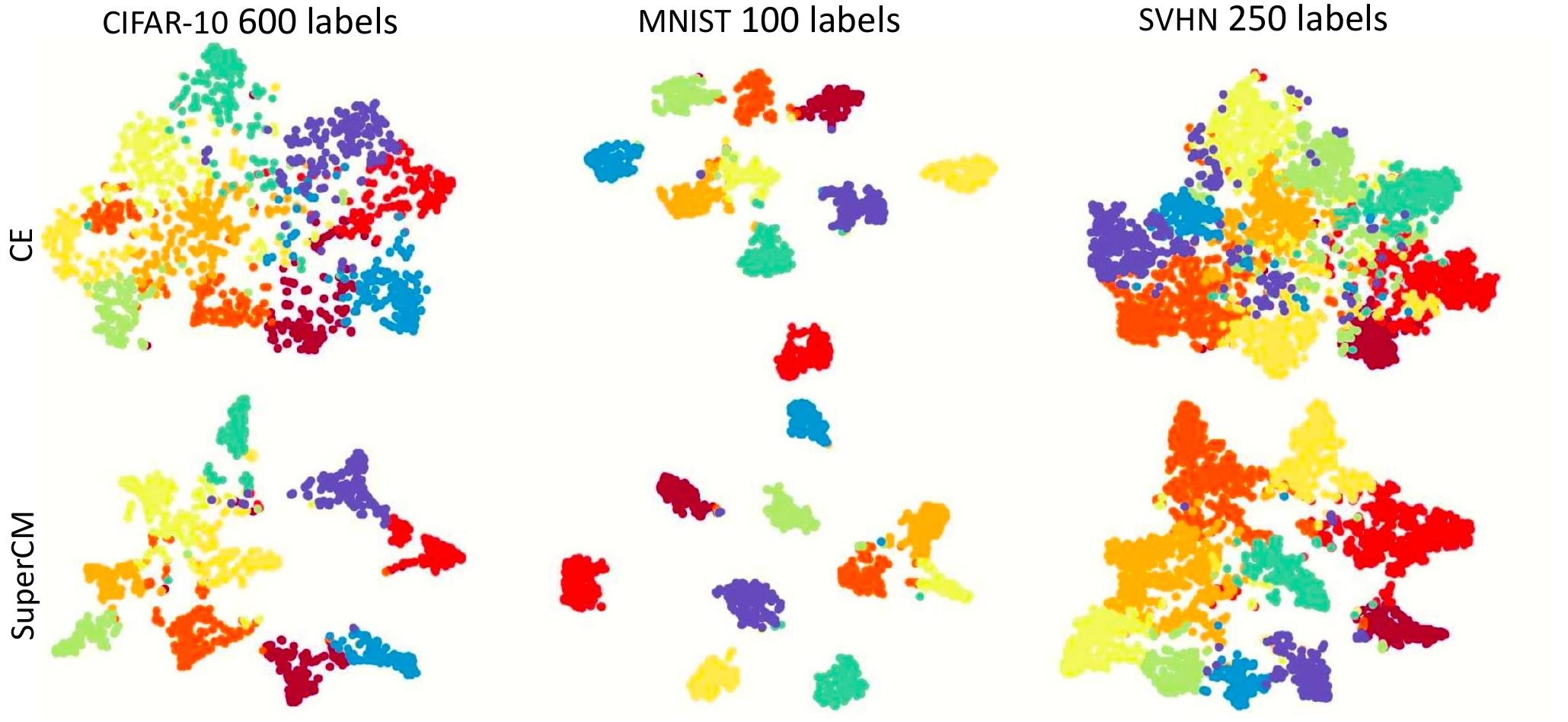}
\caption{t-SNE~\cite{tsne} plots for backbone features of the models trained with CE and SuperCM. The visualization shows different levels of supervision for CIFAR-10~\cite{cifar10-100}, MNIST~\cite{mnist}, and SVHN~\cite{svhn} datasets.}
\label{fig:umap}
\end{figure*}

We show in Figure~\ref{fig:umap} a t-SNE \cite{tsne} representations of the feature space learned by the models trained with CE and SuperCM for the CIFAR-10, SVHN, and MNIST datasets and performed experiments with CIFAR-10 600 labels, MNIST 100 labels and SVHN 250 labels settings respectively. We observe from the t-SNE plots that SuperCM training yields more separated and compact representations, which leads to better generalization and thereby reflects better performance in the classifier accuracy on the test data.

\label{sec:analysis-ssl}

\section{Unsupervised Domain Adaptation Experiments}
\label{sec:experiments-da}
We will now provide in-depth details of the UDA experiments. In this setup, SuperCM acts as an unsupervised regularizer alongside a base UDA method. We utilize DANN~\cite{Ganin2015DANN} as the base UDA approach to evaluate the performance improvements facilitated by SuperCM.

\subsection{Datasets}
\label{sec:da-dataset}
We now present a detailed overview of the public datasets used for the domain adaptation task.

\paragraph{\textbf{Office-31}}
The Office-31~\cite{office-31} dataset is widely used for domain adaptation, featuring 31 object categories across three domains: Amazon~(A), DSLR~(D), and Webcam~(W). The dataset includes 2817 images from the Amazon domain, 498 high-resolution images from DSLR cameras, and 795 low-resolution images from webcams, totaling 4110 images.

\paragraph{\textbf{Office-Home}}
The Office-Home~\cite{office-home} dataset is widely used for domain adaptation, featuring four domains Artistic (Ar), Clipart~(Cl), Product~(Pr), and Real-World~(Rw) with 65 object categories each. Artistic includes art-style images, Clipart contains clip-art style images, Product features professional product images, and Real-World includes images from real-world environments. This dataset poses a significant challenge for domain adaptation algorithms to extract robust, domain-invariant features for classification.

\paragraph{\textbf{Office-Caltech}}
The Office-Caltech~\cite{Office-caltech} dataset serves as a benchmark for domain adaptation in computer vision. It contains images from 10 overlapping categories from the Office-31~\cite{office-31} and Caltech~(C) dataset where each category contains 8 to 151 images per domain, yielding 2533 images in total.

\paragraph{\textbf{ImageClef}}
The ImageClef~\cite{image-clef14} dataset, part of the Cross-Language Evaluation Forum (CLEF) initiative, is a diverse collection sourced from Caltech-256~(C), ImageNet~(I), and Pascal VOC~(P). With 12 object categories, each comprising 50 images, ImageClef offers diversity and complexity across domains. It serves as a realistic test bed for domain adaptation tasks.

\subsection{Architecture}
In our UDA experiments, we use the Resnet-50~\cite{He2015DeepRL} architecture as the backbone along with SuperCM. The Resnet-50 architecture has 50 layers resulting in around 25 million parameters and returns feature vectors of dimension 2048.

\subsection{Experimental Settings}
This section provides implementation details of all experiments conducted in the UDA setting extensively. We provide details regarding data prepossessing, training, and evaluation protocol in the subsequent subsections.

\subsubsection{Data Preprocessing}
We conduct experiments for UDA regarding single source target setting, where the model is trained on the labeled data from the source domain and unlabeled data in the target domain in order to adapt the model to improve the source-only baseline evaluated on the target data. We performed a 70-10-20 split for training, validation, and test split for the source data and used full unlabeled data for the target domain during the training process.

\subsubsection{Baseline}
\label{subsec:baselines_uda}
In line with the SSL regularization, we performed experiments demonstrating the effectiveness of SuperCM as a regularizer in the UDA task and considered the performance of the UDA base model i.e DANN+CE~\cite{Ganin2015DANN}  as the baseline for these experiments. This baseline is a special case of Eq.~\eqref{eq:supcm} when $\delta\neq0$ and $\beta=0$.

\subsubsection{Model Training}
We adopt the training procedure outlined in~\cite{Ganin2015DANN}  for all unsupervised domain adaptation (UDA) experiments. The DANN model is trained using stochastic gradient descent with Nesterov momentum~\cite{Nesterov1983AMF}, using a momentum coefficient of $0.9$. The learning rate follows a polynomial decay schedule.

Following the approach in~\cite{Ganin2015DANN} , the domain adaptation loss weight $\delta$ is progressively increased from 0 to 1 over the course of training according to the following schedule:

\begin{equation}
\delta = \frac{2}{1 + \exp(-\gamma \cdot p)} - 1
\end{equation}

Here, $p$ denotes the training progress, linearly scaled from 0 to 1, and $\gamma = 10$ controls the steepness of the ramp-up. Additionally, the hyperparameter $\beta$ was fixed at $0.9$ for all UDA experiments where applicable.

\subsubsection{Evaluation Metrics}
We evaluate the Top-1 accuracy for the model selection and evaluate performance using training and validation data. The performance of the model is measured using Top-1 accuracy on the test set. We report mean and standard deviation over three runs, trained with different random seeds.

\setlength\LTleft{0pt}
\setlength\LTright{0pt}
\setlength\LTpre{1cm}
\setlength\LTcapwidth{\textwidth}
\begin{longtable}{@{\extracolsep{\fill}}lll}

\caption{Results for Top-1 accuracy with CE and SuperCM for DANN~\cite{Ganin2015DANN} as the base model on various datasets in UDA scenario. Bold numbers indicate statistically significant improvements (t-test, $p \leq 0.05$).} \\
\toprule
Domains                & DANN~\cite{Ganin2015DANN}+CE & DANN~\cite{Ganin2015DANN} +SuperCM\\\midrule \midrule
\multicolumn{3}{c}{Office-31~\cite{office-31}}                                         \\
A$\rightarrow$W      & 80.80$\pm$1.10       & \textbf{91.03$\pm$0.36}\\
D$\rightarrow$W      & 97.19$\pm$0.26       & 97.99$\pm$0.45\\
W$\rightarrow$A      & 65.90$\pm$1.07       & \textbf{71.91$\pm$0.25}\\
W$\rightarrow$D      & 99.67$\pm$0.23       & 99.93$\pm$0.12\\
A$\rightarrow$D      & 78.58$\pm$0.87       & \textbf{88.08$\pm$0.38}\\
D$\rightarrow$A      & 64.76$\pm$0.36       & \textbf{72.61$\pm$0.10}\\   
\textbf{Average}     & 81.15$\pm$0.65       & \textbf{86.92$\pm$0.28} \\
\midrule
\multicolumn{3}{c}{Office-Home~\cite{office-home}}                                       \\
Ar$\rightarrow$Cl    & 50.23$\pm$0.46       & \textbf{52.81$\pm$0.78}\\
Ar$\rightarrow$Pr    & 65.71$\pm$0.49       & 66.34$\pm$0.54\\
Ar$\rightarrow$Rw    & \textbf{75.10$\pm$0.17}       & 74.58$\pm$0.25\\
Cl$\rightarrow$Ar    & 54.11$\pm$0.62       & \textbf{57.24$\pm$0.28}\\
Cl$\rightarrow$Pr    & 64.58$\pm$0.88       & 64.86$\pm$0.45\\
Cl$\rightarrow$Rw    & 66.78$\pm$0.37       & 67.36$\pm$0.26\\
Pr$\rightarrow$Ar    & 53.28$\pm$0.68       & \textbf{58.32$\pm$0.68}\\
Pr$\rightarrow$Cl    & 46.65$\pm$0.32       & \textbf{50.79$\pm$0.60}\\
Pr$\rightarrow$Rw    & 74.10$\pm$0.09       & \textbf{76.89$\pm$0.48}\\
Rw$\rightarrow$Ar    & 64.65$\pm$0.83       & \textbf{68.47$\pm$0.12}\\
Rw$\rightarrow$Cl    & 52.90$\pm$0.31       & \textbf{57.38$\pm$0.49}\\
Rw$\rightarrow$Pr    & 78.41$\pm$0.10       & \textbf{80.09$\pm$0.28}\\ 
\textbf{Average}     & 62.21$\pm$0.44       & \textbf{64.59$\pm$0.43} \\
\midrule
\multicolumn{3}{c}{Office-Caltech~\cite{Office-caltech}}                               \\
A$\rightarrow$C      & 91.34$\pm$0.36       & \textbf{94.84$\pm$0.13}\\
D$\rightarrow$C      & 91.28$\pm$0.27       & \textbf{94.07$\pm$0.10}\\
W$\rightarrow$C      & 93.36$\pm$0.40       & \textbf{95.37$\pm$0.00}\\
C$\rightarrow$A      & 93.85$\pm$0.48       & 94.30$\pm$0.77\\
C$\rightarrow$D      & 82.49$\pm$1.93       & 85.23$\pm$0.73\\
C$\rightarrow$W      & 92.12$\pm$0.85       & \textbf{96.62$\pm$1.47}\\  
\textbf{Average}     & 90.74$\pm$0.71       & \textbf{93.41$\pm$0.53} \\

\midrule

\multicolumn{3}{c}{ImageClef~\cite{image-clef14}}              \\                            
I$\rightarrow$C      & 94.89$\pm$0.25       & \textbf{96.33$\pm$0.00}\\
I$\rightarrow$P      & 78.91$\pm$0.54       & 79.30$\pm$0.59\\
P$\rightarrow$C      & 93.33$\pm$0.17       & \textbf{94.61$\pm$0.38}\\
P$\rightarrow$I      & 90.33$\pm$0.00       & \textbf{92.08$\pm$0.12}\\
C$\rightarrow$I      & 93.00$\pm$0.44       & 93.39$\pm$0.48\\
C$\rightarrow$P      & 80.09$\pm$0.80       & 79.98$\pm$0.35\\ 
\textbf{Average}     & 88.42$\pm$0.37       & \textbf{89.28$\pm$0.32} \\
\bottomrule
\label{tab:results-da} 
\end{longtable}
\label{sec:results-da}

\section{ UDA Results}
Table~\ref{tab:results-da} summarizes the results for the UDA experiments regarding different source and target domain pairs on various datasets mentioned in Section~\ref{sec:da-dataset}. We discuss the effect of SuperCM training on the UDA regularization task in subsequent sections.

\paragraph{\textbf{Office-31}}

We observe in  Table~\ref{tab:results-da} that  SuperCM demonstrated effectiveness in domain adaptation between A \& W  and A \& D despite notable domain shifts. It achieves a $9.50\%$ accuracy boost in A $\rightarrow$ D and a $7.85\%$ improvement in D $\rightarrow$ A compared to CE. Notably, SuperCM consistently outperforms CE in A $\rightarrow$ W and W $\rightarrow$ A, with substantial accuracy improvements of $10.23\%$ and $6.01\%$, respectively. In cases without significant improvement, both models obtain near-perfect performance. These results highlight that clustering-based SuperCM regularization is beneficial for the UDA task.

\paragraph{\textbf{Office-Home}} 
Here, the domain shift is the least between Pr \& Rw. SuperCM notably enhances performance by $1.68\%$ for 
Rw$\rightarrow$Pr and $2.79\%$ for Pr$\rightarrow$Rw. Despite a slightly more challenging shift between Ar and Pr/Rw, SuperCM still improves performance, except for the slight decrease in Ar$\rightarrow$Rw. For the most challenging domain (Cl), SuperCM consistently improves, notably by $3.13\%$ for Cl$\rightarrow$Ar, $4.14\%$ for Pr$\rightarrow$Cl, and $4.48\%$ for Rw$\rightarrow$Cl. 

\paragraph{\textbf{Office-Caltech}}
We performed two sets of experiments when (i)~C is the target domain and (ii)~C is the source domain. For the cases with A$\rightarrow$C, D$\rightarrow$C \& W$\rightarrow$C SuperCM regularization improves the performance of the base model by $3.50\%$, $2.79\%$ and $2.01\%$, respectively. Furthermore, for the second set of experiments where C is the source domain, significant performance improvement of 4.50\%, 2.74\% is achieved in 
C$\rightarrow$W \& C$\rightarrow$D cases respectively, however, we achieve marginal performance improvement of 0.45\% for C$\rightarrow$A case.

\paragraph{\textbf{ImageClef}}
When P  is the target domain, there is a slight decrease in performance of $0.11\%$ for C$\rightarrow$P, but SuperCM shows improvement of $0.39\%$ for I$\rightarrow$P. For I, performance improves by $0.39\%$ for C$\rightarrow$I and $1.75\%$ for P$\rightarrow$I. Lastly, for C, the improvement is $1.44\%$ for I$\rightarrow$C and $1.28\%$ for P$\rightarrow$C. We can conclude from these results that SuperCM training is beneficial in the majority of cases for the UDA task on the ImageClef dataset. 

\paragraph{\textbf{Summary}}
We demonstrated that SuperCM consistently enhances the base model's performance in a majority of the domain adaptation scenarios, where domain divergence between source and target domain is high thereby benefiting the CE training with robust clustering-based regularization for the UDA task.  

\section{UDA Analysis}

\label{sec:analysis-da}
\subsection{Proxy-$\mathcal{A}$ distance}
To measure the degree of domain alignment, we conducted experiments to measure Proxy-$\mathcal{A}$ distance (PAD) which involves training an SVM classifier on backbone features to distinguish the source and target domain~\cite{BenDavid2010ATO}. We conducted these experiments on two random configurations of the Office-31 and Office-Caltech datasets. A lower PAD value corresponds to a better alignment of the source and target domain. We conducted experiments to calculate the PAD value using the backbone features trained with CE and SuperCM and the results are shown in Table~\ref{tab:pad}. We observe that training with SuperCM decreases the PAD value for different DA scenarios resulting in the alignment of source and target domain features.

\setlength\LTleft{0cm}
\setlength\LTright{0cm}
\setlength\LTpre{1cm}
\setlength\LTcapwidth{\linewidth}
\begin{longtable}{@{\extracolsep{\fill}}ccc}

    \caption{ Proxy-$\mathcal{A}$ distance(PAD)~\cite{BenDavid2010ATO} measured between each pair of domains with CE and SuperCM training for DANN~\cite{Ganin2017}. Bold numbers indicate statistically significant improvements (t-test, $p \leq 0.05$).} \\
    
    \toprule
    config & CE & SuperCM \\\midrule \midrule
    \multicolumn{3}{c}{Office-31~\cite{office-31}} \\ 
    A $\rightarrow$ D & 1.97 $\pm$ 0.02 & 1.92 $\pm$ 0.03 \\
    A $\rightarrow$ W & 1.98 $\pm$ 0.00 & \textbf{1.92 $\pm$ 0.00} \\ \midrule 
    \multicolumn{3}{c}{Office-Caltech~\cite{Office-caltech}} \\ 
    D $\rightarrow$ C & 1.88 $\pm$ 0.02 & \textbf{1.82 $\pm$ 0.02} \\
    W $\rightarrow$ C & 1.90 $\pm$ 0.01 & \textbf{1.83 $\pm$ 0.04} \\
    \bottomrule
    \label{tab:pad}
\end{longtable}

\subsection{Centroids Learning}
In this section, we conduct an ablation study to investigate different strategies to determine CM centroids within the SuperCM training framework. We explore three strategies to compute the centroids, namely ground-truth source (GS), ground-truth source+pseudo-target (GS+PT), and learned (L). In the GS strategy, the centroids are computed solely from the labeled source domain features (Eq.~\ref{eq:setcen}). For GS+PT, we extend the previous approach by including high-confidence target features ($\tau\geq 0.9$) in addition to source features for the centroids computation. Lastly, in the L strategy, the centroids are learned. 

\begin{figure}[t]
\centering
\includegraphics[width=\textwidth]{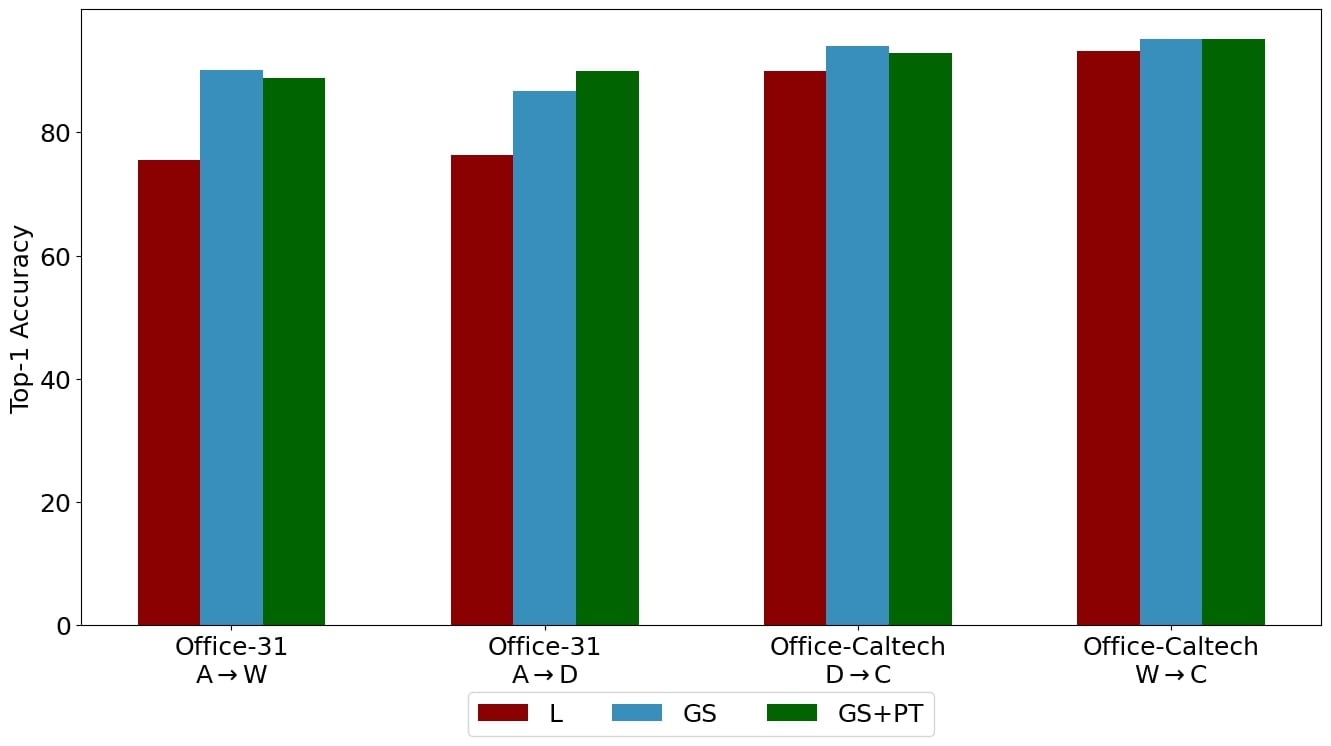}
\caption{Ablation on different DA configurations for determining centroids. The visualization shows different domain pairs for the Office-Caltech~\cite{Office-caltech} and Office-31~\cite{office-31} dataset.}
\label{fig:setcen}
\end{figure}

We observe in Figure~\ref{fig:setcen} that the GS and GS + PT strategies result in substantial performance improvements compared to the L strategy. This underscores the importance of determining centroids from labeled data for enhancing classifier performance. 
While we observe that both the GS \& GS+PT strategy improves the performance compared to the L strategy, the improvement of GS+PT over GS is less clear. Therefore, we advocate the simpler GS strategy for a reliable centroid computation.

\begin{figure*}[t]
\centering
\includegraphics[width=\textwidth]{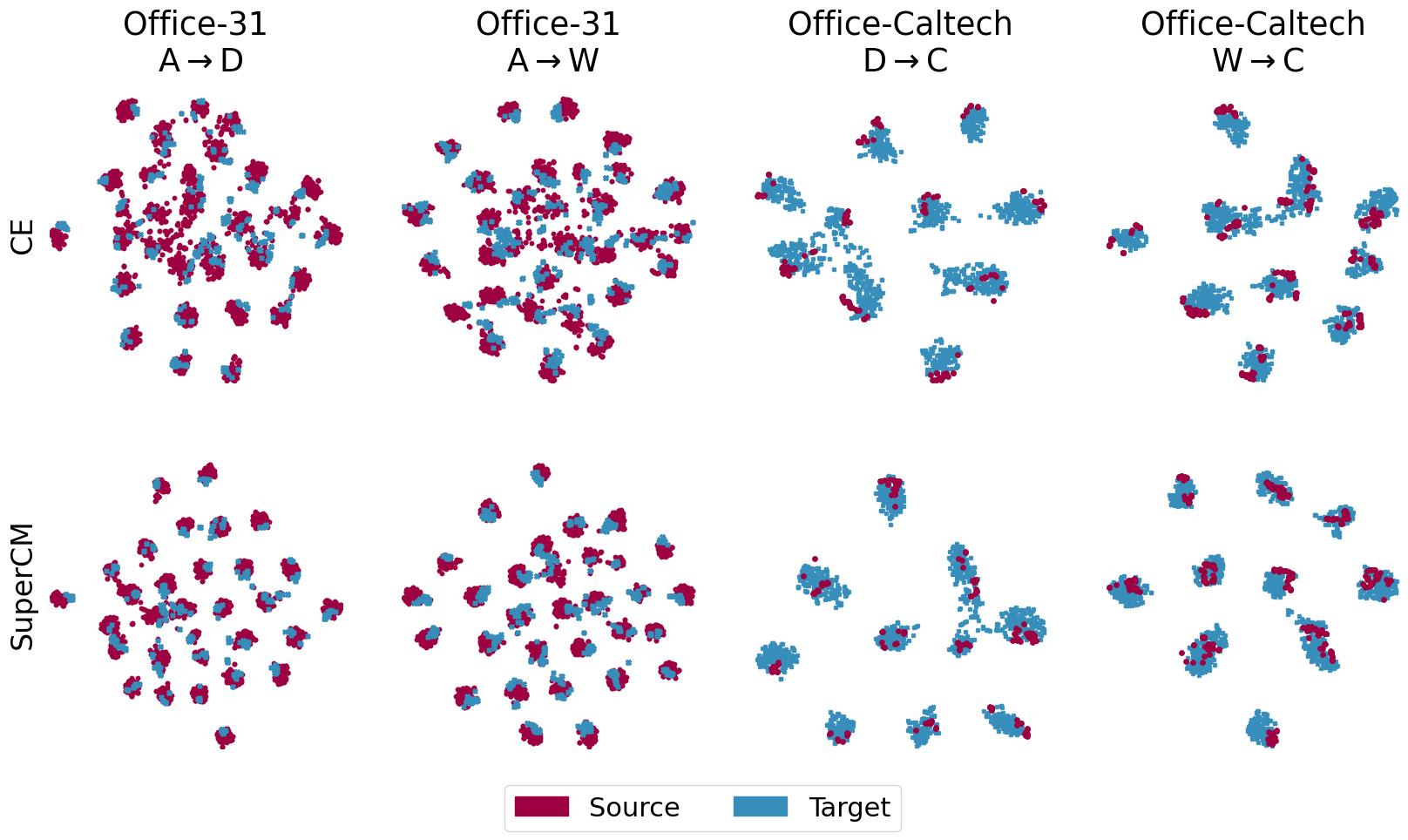}
\caption{t-SNE~\cite{tsne} visualization of backbone features during the DANN training. The visualization shows different domain pairs for the Office-Caltech~\cite{Office-caltech} and Office-31~\cite{office-31} dataset when CE and SuperCM are applied along with the base model training.}
\label{fig:umapda}
\end{figure*}

\subsection{Pre-classifier Feature Visualization}
In this section, similar to SSL experiments, we visualize the learned backbone features using t-SNE~\cite{tsne}, as depicted in Figure~\ref{fig:umapda}. The observations reveal that the representations acquired through SuperCM training, which is based on explicit clustering, exhibit improved alignment of source and target domain clusters in the feature space. This enhanced alignment helps in the unsupervised regularization of base domain adaptation training, resulting in better backbone features, and consequently improving the overall performance of the base model.

\subsection{Improving Performance of Other UDA Regularizers}
This section explores the complementary effects of SuperCM alongside other state-of-the-art UDA regularizers, specifically MCC~\cite{MCCLoss2020} and BNM~\cite{BNM2020}, for DANN~\cite{Ganin2015DANN} . We conducted experiments across various source and target domain pairs from the office datasets and showed further performance improvement when SuperCM is combined with BNM~\cite{BNM2020} and MCC~\cite{MCCLoss2020} in Table~\ref{tab:results_da2}.
The DANN+MCC model demonstrated notable improvement with the addition of SuperCM for the Office-31~\cite{office-31} dataset. Specifically, the A $\rightarrow$ W, W $\rightarrow$ A, and D $\rightarrow$ A domain adaptations showed improvements of $1.13\%$, $2.66\%$, and $1.88\%$, respectively. However, the W $\rightarrow$ D adaptation showed a marginal decline of $0.13\%$. Similarly, for the more complex Office-Home~\cite{office-home} dataset, the addition of SuperCM also led to significant gains in certain domain configurations. Notably, Pr $\rightarrow$ Ar improved by $4.97\%$, Rw $\rightarrow$ Ar by $4.87\%$, and Pr $\rightarrow$ Cl by $4.44\%$ Similarly, Cl $\rightarrow$ Ar showed a $2.71\%$ improvement.

\setlength\LTleft{0pt}
\setlength\LTright{0pt}
\setlength\LTpre{1cm}
\setlength\LTcapwidth{\textwidth}
\begin{longtable}[h!]{@{\extracolsep{\fill}}llllll}

\caption{Results demonstrate the complementary effect of SuperCM with other UDA regularizers i.e MCC~\cite{MCCLoss2020} and BNM~\cite{BNM2020}, showing Top-1 accuracy across domains on Office-31~\cite{office-31} and Office-Home~\cite{office-home} datasets. Bold numbers indicate statistically significant improvements (t-test, $p \leq 0.05$).} \\
\toprule
              & DANN~\cite{Ganin2015DANN}          & DANN~\cite{Ganin2015DANN} & DANN~\cite{Ganin2015DANN}          & DANN~\cite{Ganin2015DANN} \\ 
              & +MCC~\cite{MCCLoss2020}          & +MCC~\cite{MCCLoss2020} & +BNM~\cite{BNM2020}          & +BNM~\cite{BNM2020} \\
Domains          &           & +SuperCM &           & +SuperCM \\ 
\midrule \midrule
\multicolumn{5}{c}{Office-31~\cite{office-31}} \\ 
A$\rightarrow$W  & 91.58$\pm$1.29    & \textbf{92.71$\pm$0.47} & 92.34$\pm$0.57 & \textbf{92.34$\pm$0.77} \\ 
D$\rightarrow$W  & 98.20$\pm$0.16    & 98.45$\pm$0.16    & 98.58$\pm$0.12 & 98.32$\pm$0.21 \\ 
W$\rightarrow$A  & 72.84$\pm$0.71    & \textbf{75.50$\pm$0.63} & 72.77$\pm$0.76 & \textbf{75.6$\pm$0.56} \\ 
W$\rightarrow$D  & \textbf{99.93$\pm$0.09}    & 99.80$\pm$0.00    & 100.0$\pm$0.00 & 100.0$\pm$0.00 \\ 
A$\rightarrow$D  & 88.51$\pm$1.16    & \textbf{89.51$\pm$0.90} & 88.78$\pm$0.28 & 88.98$\pm$0.16 \\ 
D$\rightarrow$A  & 72.97$\pm$0.23    & \textbf{74.85$\pm$0.33} & 72.79$\pm$0.50 & \textbf{75.40$\pm$0.40} \\
\textbf{Average} & 87.34$\pm$0.61 & \textbf{88.47$\pm$0.42} & 87.54$\pm$0.37 & 88.44$\pm$0.35 \\
\midrule
\multicolumn{5}{c}{Office-Home~\cite{office-home}} \\ 

Ar$\rightarrow$Cl  & 56.49$\pm$0.39    & 57.70$\pm$0.53    & \textbf{50.19$\pm$0.11} & 48.66$\pm$0.23 \\ 
Ar$\rightarrow$Pr  & \textbf{72.31$\pm$0.29}    & 71.20$\pm$0.41    & \textbf{65.72$\pm$0.98} & 64.21$\pm$0.30 \\ 
Ar$\rightarrow$Rw  & 77.47$\pm$0.11    & 77.45$\pm$0.12    & \textbf{74.67$\pm$0.27} & 73.71$\pm$0.21 \\ 
Cl$\rightarrow$Ar  & 61.49$\pm$0.29    & \textbf{64.20$\pm$0.21} & 53.10$\pm$0.59 & 53.98$\pm$0.94 \\ 
Cl$\rightarrow$Pr  & 72.74$\pm$0.55    & 73.42$\pm$0.08    & 62.97$\pm$0.47 & 65.00$\pm$2.48 \\ 
Cl$\rightarrow$Rw  & 72.49$\pm$0.18    & 73.44$\pm$0.21    & 65.01$\pm$0.48 & \textbf{66.09$\pm$0.97} \\ 
Pr$\rightarrow$Ar  & 59.14$\pm$0.19    & \textbf{64.11$\pm$0.58} & 50.88$\pm$0.54 & \textbf{54.82$\pm$0.12} \\ 
Pr$\rightarrow$Cl  & 52.48$\pm$0.10    & \textbf{56.92$\pm$0.22} & 46.44$\pm$0.22 & \textbf{50.92$\pm$1.63} \\ 
Pr$\rightarrow$Rw  & 78.25$\pm$0.23    & \textbf{80.86$\pm$0.03} & 73.57$\pm$0.43 & \textbf{75.55$\pm$0.17} \\ 
Rw$\rightarrow$Ar  & 68.56$\pm$0.14    & \textbf{73.43$\pm$0.15} & 62.71$\pm$0.29 & \textbf{67.13$\pm$0.42} \\ 
Rw$\rightarrow$Cl  & 60.32$\pm$0.62    & \textbf{62.62$\pm$0.41} & 52.46$\pm$0.44 & \textbf{56.09$\pm$0.52} \\ 
Rw$\rightarrow$Pr  & 82.04$\pm$0.19    & \textbf{83.37$\pm$0.17} & 79.57$\pm$0.23 & \textbf{80.15$\pm$0.26} \\
\textbf{Average} & 67.82$\pm$0.27 & \textbf{69.89$\pm$0.26} & 61.44$\pm$0.42 & \textbf{63.03$\pm$0.69} \\

\\ 
\bottomrule
\label{tab:results_da2}
\end{longtable}

Furthermore, In the DANN+BNM model, the integration of SuperCM resulted in significant improvements across various domain adaptations in the Office-31~\cite{office-31} dataset. Specifically, the W $\rightarrow$ A and D $\rightarrow$ A adaptations demonstrated notable performance gains of $2.83\%$ and $2.61\%$, respectively. Similarly, for the Office-Home~\cite{office-home} dataset, SuperCM also led to substantial enhancements for several domain pairs. Key improvements included gains of $4.42\%$ for Rw $\rightarrow$ Ar, $4.48\%$ for Pr $\rightarrow$ Cl, and $3.63\%$ for Rw $\rightarrow$ Cl. Furthermore, the Pr $\rightarrow$ Ar adaptation saw an increase of $3.94\%$, while Cl $\rightarrow$ Rw benefited from an improvement of $1.08\%$.
In summary, the combination of SuperCM with other UDA regularizers like MCC and BNM leads to notable performance improvements across various domain adaptations. These findings underscore the potential of integrating SuperCM to strengthen existing regularizers, resulting in enhanced performance for the UDA task.

\section{Conclusion}\label{sec:conclusion}

In this paper, we have demonstrated that the proposed SuperCM framework substantially improves upon existing baselines in both SSL and UDA tasks by incorporating an explicit clustering-based regularization strategy inspired by GMMs. By leveraging the clustering assumption, SuperCM enhances the quality of feature representations and decision boundaries.

In the SSL setting, SuperCM consistently improves the performance of base models across different levels of supervision. This highlights its adaptability and effectiveness in low-label regimes, which is particularly valuable in real-world scenarios where annotated data is limited. Furthermore, SuperCM integrates seamlessly with diverse SSL backbones, contributing to notable gains through its structured regularization of the feature space. In the UDA setting, SuperCM acts as an effective regularizer for improving alignment between source and target domains. This is evidenced by quantitative improvements in Proxy-$\mathcal{A}$ distance metrics as well as qualitative enhancements observed in t-SNE visualizations, which demonstrate better inter-class separation and intra-class compactness. These improvements underscore the robustness and generalizability of SuperCM across both domains and datasets.

While SuperCM demonstrates substantial gains in SSL setting, we observe that the magnitude of improvement tends to diminish as the level of supervision increases. This suggests that hybrid strategies—such as combining SuperCM with complementary SSL techniques that could reinforce its effectiveness in high-supervision regimes—may offer additional benefits. Additionally, few-shot learning remains a promising avenue for extending SuperCM, particularly in scenarios with limited labeled data. Its integration into self-supervised learning pipelines could further enhance generalization under minimal supervision, which is especially relevant in domains like medical imaging, where annotated data is often scarce and costly. As the volume of data and the complexity of tasks continue to grow, the scalability of SuperCM becomes increasingly important. Future work could explore how SuperCM's performance and computational efficiency scale with larger datasets and more complex models, providing deeper insights into its practical applicability in real-world settings.

Beyond vision tasks, SuperCM could extend to domains like text or tabular data, where it has potential for robust feature learning in settings with limited or noisy labels—e.g., clustering semantically similar phrases in sentiment analysis or grouping patient profiles in clinical datasets. In more complex domain adaptation scenarios—such as multi-source or open-set transfer—SuperCM could support target-domain learning by leveraging cluster structures from source domains. Future extensions could explore the benefits of SuperCM regularization for complex scenarios of domain adaptation.

In summary, SuperCM provides a robust and flexible framework that enhances SSL and UDA performance through explicit, class-conditional clustering regularization and it shows strong potential for broader use across domains.

\end{document}